\documentclass[11pt,a4paper]{article}

\usepackage[utf8]{inputenc}
\usepackage[T1]{fontenc}
\usepackage{lmodern}
\usepackage[margin=1in]{geometry}
\usepackage{authblk}
\usepackage{hyperref}
\usepackage{booktabs}
\usepackage{listings}
\usepackage{xcolor}
\usepackage{natbib}
\usepackage{parskip}
\usepackage{graphicx}

\definecolor{codegreen}{rgb}{0,0.6,0}
\definecolor{codegray}{rgb}{0.5,0.5,0.5}
\definecolor{codepurple}{rgb}{0.58,0,0.82}
\definecolor{backcolour}{rgb}{0.95,0.95,0.92}

\lstdefinestyle{mystyle}{
    backgroundcolor=\color{backcolour},   
    commentstyle=\color{codegreen},
    keywordstyle=\color{magenta},
    numberstyle=\tiny\color{codegray},
    stringstyle=\color{codepurple},
    basicstyle=\ttfamily\footnotesize,
    breakatwhitespace=false,         
    breaklines=true,                 
    captionpos=b,                    
    keepspaces=true,                 
    showspaces=false,                
    showstringspaces=false,
    showtabs=false,                  
    tabsize=4
}
\hypersetup{
    colorlinks=true,
    linkcolor=blue,
    filecolor=magenta,      
    urlcolor=cyan,
    citecolor=blue,
}

\title{\textbf{\texttt{bde}: A Python Package for Bayesian\\Deep Ensembles via MILE}}

\author[1,*]{Vyron Arvanitis}
\author[1,*]{Angelos Aslanidis}
\author[2,3,*]{Emanuel Sommer}
\author[2,3]{David R\"ugamer}

\affil[1]{Faculty of Physics, LMU Munich, Munich, Germany}
\affil[2]{Department of Statistics, LMU Munich, Munich, Germany}
\affil[3]{Munich Center for Machine Learning, Munich, Germany}
\affil[*]{These authors contributed equally.}
\affil[ ]{Correspondence to \texttt{emanuel.sommer@stat.uni-muenchen.de}}

\date{\today}

\begin{document}

\maketitle

\section*{Summary}

\texttt{bde} (\href{https://github.com/scikit-learn-contrib/bde}{https://github.com/scikit-learn-contrib/bde}) is a Python package designed to bring state-of-the-art sampling-based Bayesian Deep Learning (BDL) to practitioners and researchers. The package combines the speed and high-performance capabilities of JAX and \texttt{blackjax} \citep{jax2018github, cabezas2024blackjax} with the user-friendly API of scikit-learn \citep{scikit-learn}. It targets tabular supervised learning tasks, including distributional regression and (multi-class) classification, providing a seamless interface for Bayesian Deep Ensembles (BDEs) \citep{sommer2024connecting} specifically implementing \textbf{Microcanonical Langevin Ensembles (MILE)} \citep{sommer2025mile}.

The workflow of \texttt{bde} implements the two-stage BDE inference process of MILE. First, it optimizes \texttt{n\_members} independent instances of a flexibly configurable feed-forward neural network using regularized empirical risk minimization (with the negative log-likelihood as loss) via the AdamW optimizer \citep{loshchilov2018decoupled}. Second, it transitions to a sampling phase using Microcanonical Langevin Monte Carlo \citep{robnik2023microcanonical, robnik2024fluctuation}, enhanced with a tuning phase adapted for Bayesian Neural Networks. This combination is referred to as MILE \citep{sommer2025mile}. In essence, the optimization of the ensemble of neural networks first finds diverse high-likelihood modes, from which sampling then explores local posterior structure. This process generates an ensemble of samples (models) that constitute an implicit posterior approximation.

\section*{Software design}

Because optimization and sampling across ensemble members are independent, \texttt{bde} exploits JAX’s parallelization and just-in-time compilation to scale efficiently across CPUs, GPUs, and TPUs. Given new test data, the package approximates the posterior predictive, enabling point predictions, credible intervals, coverage estimates, and other uncertainty metrics through a unified interface.

\section*{State of the field}

Reliable uncertainty quantification (UQ) is increasingly viewed as a critical component of modern machine learning systems, and BDL provides a principled framework for achieving it \citep{papamarkou2024position}. While several libraries support optimization-based approaches such as variational inference or classical Bayesian modeling, accessible tools for sampling-based inference in Bayesian neural networks remain scarce. Existing probabilistic programming frameworks offer MCMC but require substantial manual configuration to achieve competitive performance on neural network models.

\section*{Statement of need}

\texttt{bde} addresses the gap outlined above by providing the first user-friendly implementation of MILE - a hybrid sampling technique shown to deliver strong predictive accuracy and calibrated uncertainty for Bayesian neural networks \citep{sommer2025mile}. By providing full scikit-learn compatibility, the package enables seamless integration into existing machine learning workflows, allowing users to obtain principled Bayesian uncertainty estimates without specialized knowledge of MCMC dynamics, initialization strategies, or JAX internals.

Through automated orchestration of optimization, sampling, parallelization, and predictive inference, \texttt{bde} offers a fast, reproducible, and practical solution for applying sampling-based BDL methods to tabular supervised learning tasks.

\section*{Research impact statement}

\texttt{bde} bridges the gap between high-performance MCMC research and practical data science. By providing a curated implementation of MILE for tabular data, it enables researchers and practitioners alike to easily apply BDEs. Its inclusion in the \texttt{scikit-learn-contrib} organization ensures adherence to rigorous software standards and API consistency, making it a trusted, community-ready tool for reproducible UQ in tabular machine learning.

\section*{Usage example}

The following example illustrates a regression task with UQ using \texttt{bde} in only a few lines of code. Training inputs are assumed to be preprocessed and normalized. The workflow specifies the ensemble and sampling hyperparameters, fits the model, and obtains posterior predictive quantities, including predictive moments, credible intervals, and raw ensemble outputs.

\begin{lstlisting}[language=Python]
from bde import BdeRegressor

regressor = BdeRegressor(
        n_members=8, # parallelizes over the available cores
        hidden_layers=[16, 16], # 2 hidden layers of width 16
        activation="relu", # (default)
        epochs=1000,
        validation_split=0.15,
        lr=1e-3,
        weight_decay=1e-4,
        patience=20,
        warmup_steps=5000,
        n_samples=200, # 200/10 x 8 = 160 final posterior samples
        n_thinning=10,
        # Controls MCMC exploration (default):
        desired_energy_var_start=0.5, 
        desired_energy_var_end=0.1,
)

regressor.fit(x=X_train, y=y_train)

means, sigmas = regressor.predict(X_test, mean_and_std=True)
means, intervals = regressor.predict(X_test, credible_intervals=[0.1, 0.9])
raw = regressor.predict(X_test, raw=True)
\end{lstlisting}

Classification follows analogously using \texttt{BdeClassifier}.

\section*{Regression benchmark}

We provide a small benchmark of \texttt{bde} on the \texttt{airfoil} \citep{Dua_2019} and the \texttt{bikesharing} \citep{misc_bike_sharing_dataset_275} datasets. We report mean predictive performance (RMSE), UQ metrics (NLL in the distributional and mean regression formulation), reported as mean $\pm$ standard deviation over 5 independent runs. The results show competitive out-of-the-box performance of BDE especially in UQ with its native distributional regression capability.

\begin{table}[h]
\centering
\resizebox{.75\textwidth}{!}{
\begin{tabular}{lccc}
\toprule
\textbf{\texttt{airfoil}} & \textbf{RMSE} & \textbf{NLL (distr. regr.)} & \textbf{NLL (mean regr.)} \\
\midrule
Linear Model  & 0.6598 $\pm$ 0.0000 & - & 1.0032 $\pm$ 0.0000 \\
Random Forest & 0.2560 $\pm$ 0.0015 & - & 0.0567 $\pm$ 0.0057 \\
XGBoost       & 0.2025 $\pm$ 0.0055 & - & -0.1782 $\pm$ 0.0269 \\
CatBoost      & 0.2393 $\pm$ 0.0036 & 0.1479 $\pm$ 0.1414 & -0.0109 $\pm$ 0.0152 \\
Deep Ensemble & 0.3650 $\pm$ 0.0038 & 0.1760 $\pm$ 0.0043 & 0.4100 $\pm$ 0.0105 \\
TabPFN (V2)   & 0.1359 $\pm$ 0.0028 & \textbf{-0.9338 $\pm$ 0.0195} & -0.5769 $\pm$ 0.0205 \\
BDE           & \textbf{0.1215 $\pm$ 0.0042} & \textbf{-0.9126 $\pm$ 0.0131} & \textbf{-0.6888 $\pm$ 0.0347} \\
\bottomrule
\end{tabular}
}
\end{table}

\begin{table}[h]
\centering
\resizebox{.75\textwidth}{!}{
\begin{tabular}{lccc}
\toprule
\textbf{\texttt{bikesharing}} & \textbf{RMSE} & \textbf{NLL (distr. regr.)} & \textbf{NLL (mean regr.)} \\
\midrule
Linear Model  & 0.7796 $\pm$ 0.0000 & - & 1.1700 $\pm$ 0.0000 \\
Random Forest & 0.2440 $\pm$ 0.0020 & - & 0.0085 $\pm$ 0.0009 \\
XGBoost       & 0.2143 $\pm$ 0.0010 & - & -0.1215 $\pm$ 0.0049 \\
CatBoost      & 0.2652 $\pm$ 0.0021 & -0.3737 $\pm$ 0.0229 & 0.0918 $\pm$ 0.0080 \\
Deep Ensemble & 0.2880 $\pm$ 0.0052 & -0.4550 $\pm$ 0.0173 & 0.1740 $\pm$ 0.0181 \\
TabPFN (V2)   & \textbf{0.2103 $\pm$ 0.0008} & -0.6856 $\pm$ 0.0063 & \textbf{-0.1400 $\pm$ 0.0041} \\
BDE           & 0.2261 $\pm$ 0.0016 & \textbf{-0.7315 $\pm$ 0.0098} & -0.0679 $\pm$ 0.0071 \\
\bottomrule
\end{tabular}}
\end{table}

All models used 10 CPU cores without additional tuning. For BDE, we generated 1000 posterior samples from a feed-forward neural network with four hidden layers of width 16. While even highly optimized BDEs generally incur a higher computational cost than optimization-based competitors due to the iterative nature of MCMC sampling, this investment enables the construction of a flexible, non-parametric posterior approximation. This trade-off, as demonstrated above, yields strong performance and rigorous epistemic UQ, e.g. through calibrated credible intervals. All experimental configurations are shipped in the released package to ensure reproducibility.

\section*{AI usage disclosure}

Generative AI was used via GitHub Copilot for local, line- or block-level code autocompletion during software development, using the Claude Sonnet 3.7 and 4 models. Codex was additionally used to assist with the initial draft of the package documentation and minor refactors to ensure compatibility with the \texttt{scikit-learn} API. No AI tools were used for ideation, architectural or design decisions, code review, or testing strategy. All AI-generated suggestions were critically reviewed, modified where necessary, and fully validated by the authors, who retain complete responsibility.

\section*{Acknowledgments}

DR’s research is funded by the Deutsche Forschungsgemeinschaft (DFG, German Research Foundation) – 578966082.

\bibliographystyle{plainnat}
\bibliography{paper}

\end{document}